\documentclass[runningheads]{llncs}
\usepackage[T1]{fontenc}
\usepackage{graphicx}
\usepackage{booktabs}
\usepackage{epsfig}
\usepackage{amsmath}
\usepackage{amssymb}
\usepackage{booktabs}
\usepackage{xcolor,colortbl}
\usepackage{multirow}
\usepackage{subcaption}
\usepackage{hyperref}
\usepackage[misc,geometry]{ifsym}
\usepackage{makeidx} 

\usepackage{color}

\definecolor{mygray}{gray}{0.9}

\makeindex 

\begin{document}

\title{Segmenting Small Stroke Lesions with Novel Labeling Strategies}

\author{
Liang Shang
\and
Zhengyang Lou
\and
Andrew L. Alexander
\and
Vivek Prabhakaran
\and
William A. Sethares
\and
Veena A. Nair
\and
Nagesh Adluru$^{(\text{\Letter})}$
}
\authorrunning{L.Shang et al.}
%
\institute{University of Wisconsin-Madison, Madison, WI 53706, USA\\ \email{adluru@wisc.edu}}

\maketitle              
\begin{abstract}
Deep neural networks have demonstrated exceptional efficacy in stroke lesion segmentation. However, the delineation of small lesions, critical for stroke diagnosis, remains a challenge. In this study, we propose two straightforward yet powerful approaches that can be seamlessly integrated into a variety of networks: \textbf{\underline{M}}ulti-\textbf{\underline{S}}ize \textbf{\underline{L}}abeling (MSL) and \textbf{\underline{D}}istance-\textbf{\underline{B}}ased \textbf{\underline{L}}abeling (DBL), with the aim of enhancing the segmentation accuracy of small lesions. MSL divides lesion masks into various categories based on lesion volume while DBL emphasizes the lesion boundaries. 
Experimental evaluations on the Anatomical Tracings of Lesions After Stroke (ATLAS) v2.0 dataset showcase that an ensemble of MSL and DBL achieves consistently better or equal performance on recall (\textbf{3.6\%} and \textbf{3.7\%}), F1 (\textbf{2.4\%} and \textbf{1.5\%}), and Dice scores (\textbf{1.3\%} and 0.0\%) compared to the top-1 winner of the 2022 MICCAI ATLAS Challenge on both the subset only containing small lesions and the entire dataset, respectively.
Notably, on the mini-lesion subset, a single MSL model surpasses the previous best ensemble strategy, with enhancements of \textbf{1.0}\% and \textbf{0.3}\% on F1 and Dice scores, respectively.
Our code is available at: \url{https://github.com/nadluru/StrokeLesSeg}.

\end{abstract}

\section{Introduction}

Strokes, characterized by an inadequate blood supply to specific brain regions, rank as the second leading cause of death and third leading cause of disability globally~\cite{stroke_death}. Accurate segmentation of stroke lesions helps clinicians to better diagnose and evaluate any treatment risks~\cite{tsai2014automated}. For instance, segmentation aids in identifying the location and extent of the ischemic core and penumbra, which are crucial for determining the eligibility and efficacy of thrombolysis or thrombectomy~\cite{borst2015effect}. Additionally, segmentation can help monitor the evolution of the lesion over time and assess the response to treatment.

\begin{figure*}[!t]
    \centering
    \hspace{-4mm}
    \includegraphics[width=\linewidth]
    {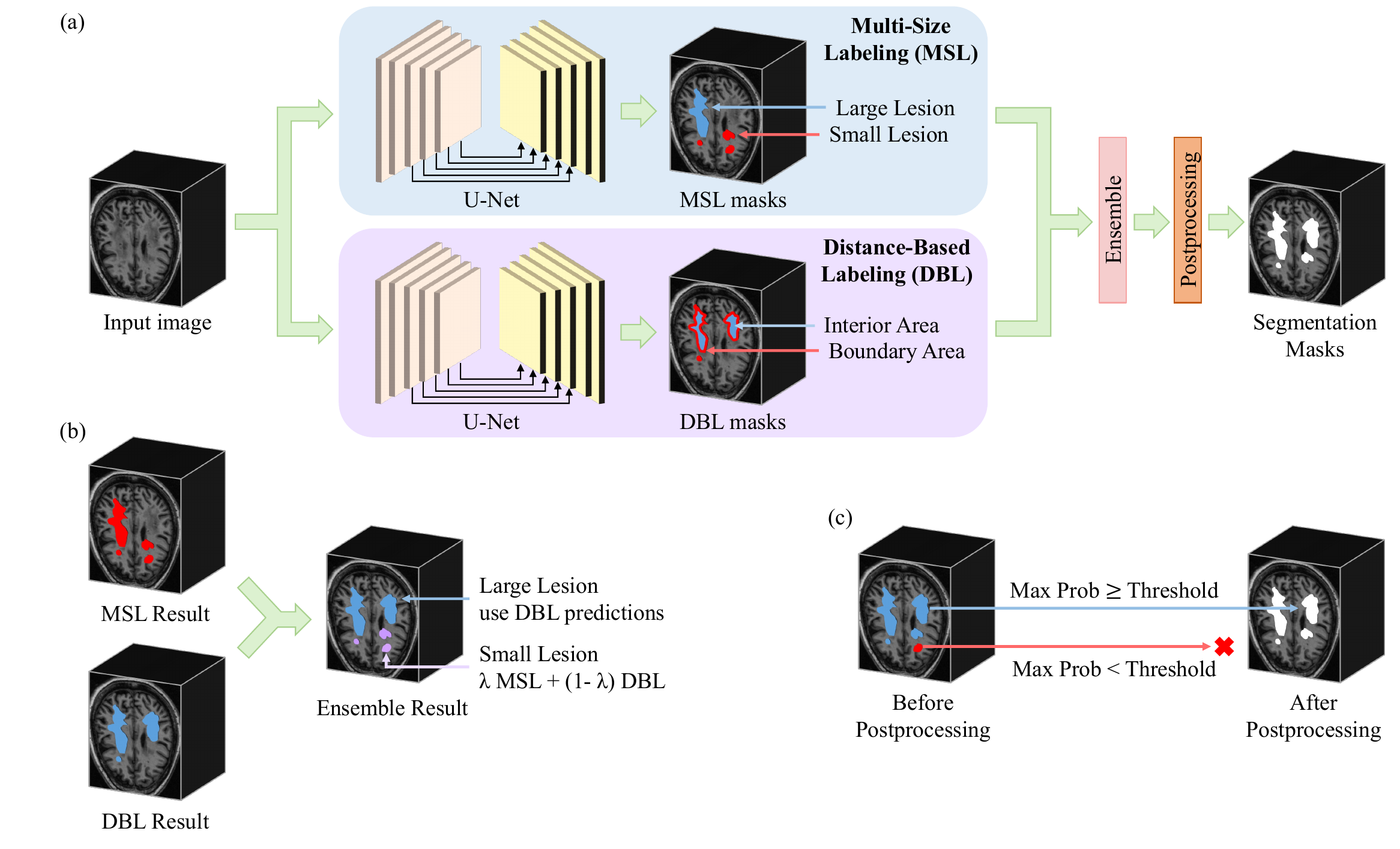}
    \caption{\textbf{(a) Overview of our method.} We first propose \textbf{\underline{M}}ulti-\textbf{\underline{S}}ize \textbf{\underline{L}}abeling (MSL) and \textbf{\underline{D}}istance-\textbf{\underline{B}}ased \textbf{\underline{L}}abeling (DBL). 
    These two labeling strategies accentuate small lesions and lesion boundaries, respectively. Then, we ensemble models trained with MSL and DBL. Following, post-processing is applied to generate the final segmentation mask.
    \textbf{(b) Proposed ensemble strategy.}
    A linear interpolation between MSL and DBL results is applied for small lesions to generate ensemble results. For large lesions, we exclusively rely on DBL results.
    \textbf{(c) Proposed postprocessing strategy.} 
    The postprocessing pipeline is designed to enhance segmentation accuracy by filtering out small lesions whose maximum probability falls below a predefined threshold.
    }
    \label{fig:overview}
\end{figure*}

The segmentation of small stroke lesions has even more significant clinical implications. Accurate segmentation and quantification of them can provide valuable information for stroke diagnosis and prognosis and for reducing the risk of progressing to post-stroke 
vascular contributions to cognitive impairment and dementia (VCID)~\cite{wardlaw2013neuroimaging,corriveau2016science}. However, small stroke lesions are often overlooked or misdiagnosed by radiologists due to their low contrast and subtle appearance. It is thus essential to develop methods that can accurately segment small lesions to assist the radiologists.  

While deep neural networks have demonstrated impressive performance in detecting~\cite{stroke_detection} and segmenting~\cite{stroke_segmentation} stroke lesions from Magnetic Resonance Imaging (MRI) scans, they still fail short in segmenting small lesions. We argue that this is because most methods assess their performance based on the match between the number of voxels correctly/incorrectly detected. Since large-volume lesions contain many more voxels than small ones, measures such as the overall Dice score are inherently biased towards accurately detecting large lesions. Accordingly, delineating small lesions remains challenging, despite their critical importance for stroke diagnosis. 

Among the existing literature, only a handful of studies focus on segmenting small brain lesions exclusively from brain MRIs. For example, \cite{small_lesion_ensemble,small_lesion_ensemble_3,small_lesion_ensemble_4} develop ensemble learning strategies that combine the outputs of multiple CNN blocks operating at different resolutions, which often require specific model architectures. SPiN~\cite{spin} explores the use of subpixel embedding to retain fine-grained details of the input, particularly in segmenting 2D MRI slices, and its application to 3D MRIs is limited due to the computational complexity associated with image super-resolution techniques.

To this end, we introduce two innovative labeling strategies, \textbf{\underline{M}}ulti-\textbf{\underline{S}}ize \textbf{\underline{L}}abeling (MSL) and \textbf{\underline{D}}istance-\textbf{\underline{B}}ased \textbf{\underline{L}}abeling (DBL), designed to categorize the binary segmentation mask into multiple classes based on the lesion volume and distance to the non-lesion region, respectively. To the best of our knowledge, our study is the first to address the often overlooked and clinically important problem of small lesion segmentation from a data augmentation perspective. Implemented without modifying the feature extractor in the U-Net~\cite{unet} architecture, MSL and DBL empower the network to differentiate between lesion types with distinct features, thus emphasizing small lesions and the boundary region, respectively. Our contributions can be summarized as follows:
\begin{itemize}
\item Our novel labeling strategies (MSL and DBL) are tailored to enhance the performance of small stroke lesion segmentation. Both strategies can be seamlessly integrated into a wide range of segmentation models.
\item We propose supplementary ensemble and postprocessing strategies that effectively leverage the advantages of MSL and DBL.
\item On the Anatomical Tracings of Lesions After Stroke (ATLAS) v2.0 dataset~\cite{ATLAS}, our ensemble strategy consistently outperforms the leading baseline ensemble in terms of recall, F1, and Dice scores across both the subset specifically targeting small lesions and the entire dataset. Moreover, a single MSL model surpasses the best baseline ensemble on the mini-lesion subset. 
\end{itemize}

\section{Approach}

Typical state-of-the-art segmentation studies on stroke lesions~\cite{mapping,spin,small_lesion_ensemble,small_lesion_ensemble_3,small_lesion_ensemble_4} employ binary segmentation masks that distinguish only between normal and lesion regions. We note that due to the inherent imbalance in lesion sizes, it may be advantageous to categorize lesions explicitly based on either their volume or their proximity to the non-lesion region. While this approach maintains the same feature extractor architecture as standard models, it encourages the final classifier to discern different lesion types based on these distinct features. To this end, we introduce \textbf{\underline{M}}ulti-\textbf{\underline{S}}ize \textbf{\underline{L}}abeling (MSL) and \textbf{\underline{D}}istance-\textbf{\underline{B}}ased \textbf{\underline{L}}abeling (DBL), aimed at enhancing the learning process for small lesion. These methods transform the voxel-wise binary classification task for stroke lesion segmentation into a voxel-wise multi-way classification task, facilitating more nuanced and effective segmentation but only adding 0.05\% additional FLOPs and 0.007\% additional parameters to the network. The segmentation pipeline of our methods is illustrated in Fig.~\ref{fig:overview}.

\subsection{Proposed Label Strategies}

\noindent
\textbf{Multi-Size Labeling (MSL).} Segmenting small lesions poses a challenge due to the extreme volume imbalance of lesions of different sizes. In the Anatomical Tracings of Lesions After Stroke (ATLAS) v2.0 dataset~\cite{ATLAS}, for instance, among 655 training images, 517 lesions have a volume of less than 100 voxels, totaling approximately 50,000 voxels. On the other hand, a single larger lesion in the dataset can exceed 100,000 voxels. Consequently, disregarding small lesions in predictions does not significantly impact measures of segmentation quality, such as the Dice score. 
To emphasize the segmentation of small lesions, we introduce \textbf{\underline{M}}ulti-\textbf{\underline{S}}ize \textbf{\underline{L}}abeling (MSL), which categorizes lesion voxels based on their volumes. Specifically, for a voxel $k$ belonging to a lesion $K$, we assign it to one of the following categories
\begin{equation}
    \text{lesion voxel } k \in \left\{
    \begin{aligned}
        &\text{tiny lesion}, & \text{if} &\ |K|<100, \\
        &\text{small lesion}, & \text{if} &\ 100\le|K|<1,000, \\
        &\text{medium lesion}, & \text{if} &\ 1,000\le|K|<10,000, \\
        &\text{large lesion}, & \text{if} &\ |K|\ge10,000,
    \end{aligned}
    \right.
\end{equation}
where $|K|$ represents the volume of the lesion. While balancing the number of voxels for MSL is not feasible due to the extreme variability in lesion sizes, employing this logarithmic categorization ensures each class contains a roughly equal number of lesions, \textit{i.e.},  within the same order of magnitude. We also explore alternative categorizations in our ablation studies.

\noindent
\textbf{Distance-Based Labeling (DBL).}
Interior regions within lesions and normal tissues tend to lack important features for segmentation. In contrast, boundary voxels may contain important information as they represent edges between the lesion and non-lesion regions.
Hence, an alternative labeling strategy aims to delineate boundaries from interior regions. We call this strategy \textbf{\underline{D}}istance-\textbf{\underline{B}}ased \textbf{\underline{L}}abeling (DBL). DBL categorizes lesion voxels $k$ as
\begin{equation}
    \text{lesion voxel } k \in \left\{
    \begin{aligned}
        &\text{boundary region}, & \text{if} &\ \text{distance to non-lesion region}\le2, \\
        &\text{interior region}, & \text{if} &\ \text{distance to non-lesion region}>2,
    \end{aligned}
    \right.
\end{equation}
to ensure a relatively narrow boundary region with the number of voxels in each class remains within the same order of magnitude. Several alternative categorizations of MSL and DBL are explored in our ablation studies. For completeness of this work, the detailed lesion distribution across categories for both MSL and DBL are listed in Sec. A of the supplementary materials.

\noindent
\textbf{Generating binary segmentation masks.} As the binary segmentation task transitions into a multi-class voxel-level classification task, additional steps are required to derive a binary segmentation mask from the output. Generally, the network logits are fed into a softmax function to compute the predicted probabilities $p_{k,i}$ of each voxel $k$ belonging to each class $i \in \{0, 1, 2...\}$. In our case, the class $0$ represents the background class, and the remaining classes denote the foreground classes assigned by MSL or DBL. The likelihood of a voxel $k$ being part of a lesion, $p_{k}$, can be obtained by aggregating the probabilities of all foreground classes by $p_{k} = \sum_{i\ge1}p_{k,i}$.
Afterwards, the segmentation mask can be achieved by thresholding
all voxels with $p_{k} > 0.5$.

\subsection{Ensemble and Postprocessing Pipelines}

\noindent
\textbf{Ensemble.}
While MSL significantly enhances the network’s capability to segment small lesions, its performance in segmenting larger lesions may be compromised due to the inherent complexity of the multi-class classification task. Similarly, although DBL adeptly highlights the boundaries of both small and large lesions, 
its efficacy may diminish when segmenting tiny lesions as their boundary and interior regions can be hard to distinguish.
To circumvent these limitations, we propose a novel ensemble strategy that integrates both MSL and DBL to ensure a balanced enhancement in the accuracy of lesion segmentation across a diverse range of sizes, thereby addressing a critical gap in current segmentation methodologies. 
This approach first calculates the probabilities of a voxel $k$ being a lesion separately from the MSL and DBL models, denoted $p_{k,\text{MSL}}$ and $p_{k,\text{DBL}}$, respectively. Then, as illustrated in Fig.~\ref{fig:overview}b, for a lesion prediction $K$ identified as tiny or small, \textit{i.e.}, $|K|<1000$, we linearly interpolate between $p_{k,\text{MSL}}$ and $p_{k,\text{DBL}}$. Conversely, for larger lesions, we rely solely on $p_{k,\text{DBL}}$. Consequently, the final likelihood of a voxel $k$ being part of a lesion is
\begin{equation}
    p_{k} = \left\{
    \begin{aligned}
        &\lambda \cdot p_{k,\text{MSL}} + (1 - \lambda) \cdot p_{k,\text{DBL}}, & \text{if} \ |K| < 1&000, \\
        &p_{k,\text{DBL}}, & \text{otherwise} & ,
    \end{aligned}
    \right.
    \label{eq:ensemble}
\end{equation}
where $\lambda$ is the mixing rate determined through ablation studies.

\noindent
\textbf{Postprocessing (PP).}
The postprocessing (PP) pipeline aims to further reduce segmentation errors for small lesions. 
As shown in Fig.~\ref{fig:overview}c, for each prediction of a tiny or small lesion $K$ with $|K| < 1000$, if the maximum probability among all lesion voxels, $\max_{k \in K} p_{k}$, is less than a pre-defined probability threshold $p_t$ determined through ablation studies, we exclude it from the segmentation mask. 

\section{Experimental Setting}

\noindent
\textbf{Dataset.} We demonstrate the effectiveness of our proposed labeling strategies using the Anatomical Tracings of Lesions After Stroke (ATLAS) v2.0 dataset~\cite{ATLAS}. 
This dataset comprises several training and testing subsets, with a training subset of 655 T\(_1\)-weighted MRIs are public with corresponding lesion segmentation masks. All the images in the dataset were corrected for intensity non-uniformity, intensity standardized and linearly registered to the MNI space, where each voxel corresponds to a volume of $1\ \text{mm}^3$.
Within the 655 MRI scans in the training dataset, the lesion volumes span from 13 to over 200,000 voxels, with 138 MRI scans exclusively containing lesions with volumes less than 1000, constituting a subset of mini-lesions.

\noindent
\textbf{Implementation.} Our models are developed using PyTorch~\cite{pytorch} and nnU-Net~\cite{nnunet}. The models are trained for 1000 epochs, with Dice together with cross-entropy (CE) as the compound loss, a batch size of 2, SGD as the optimizer, an initial learning rate of 0.01, a momentum of 0.99, and z-score normalization.
During training, 2 image patches of dimensions $128 \times 128 \times 128$ are extracted from each of the two MRI images in the batch to serve as inputs to the model. One patch is randomly cropped from the original MRI, while the other is specifically cropped to ensure that lesion voxels are centered within the patch. This approach balances the diversity of training samples with a focus on the critical regions containing lesions. 
Across all training schemes, we conducted a size-balanced 5-fold cross-validation using the data split outlined in~\cite{mapping}, whereby the training dataset is evenly partitioned into 5 folds, ensuring nearly identical lesion volume distributions within each fold. Our results are summarized across the 5-fold models over both the entire dataset (655 scans) and the mini-lesion subset (138 scans).
For evaluation. we employed four metrics, Dice, F1, precision, and recall scores, to evaluate performance. The Dice score is computed \textbf{voxel-wise}, while the latter three scores are computed \textbf{lesion-wise}. 

\section{Results}

\begin{table}[!t]\small
    \centering
    \scalebox{0.9}{
    \begin{tabular}{l|cccc|cccc}
        \toprule
        \multirow{2}{*}{Method}
        & \multicolumn{4}{c|}{Mini-Lesion Subset}
        & \multicolumn{4}{c}{Entire Dataset}
        \\
        \cline{2-9}
        & \makebox[0.084\linewidth][c]{Dice}
        & \makebox[0.084\linewidth][c]{F1}
        & \makebox[0.084\linewidth][c]{Precision}
        & \makebox[0.084\linewidth][c]{Recall}
        & \makebox[0.084\linewidth][c]{Dice}
        & \makebox[0.084\linewidth][c]{F1}
        & \makebox[0.084\linewidth][c]{Precision}
        & \makebox[0.084\linewidth][c]{Recall}
        \\
        \midrule
        Default
        & 0.417
        & 0.500
        & 0.509
        & 0.628
        & 0.635
        & 0.549
        & 0.686
        & 0.561
        \\
        DTK10
        & 0.416
        & 0.491
        & 0.550
        & 0.597
        & 0.629
        & 0.547
        & 0.697
        & 0.560
        \\
        Res U-Net
        & 0.433
        & 0.502
        & 0.481
        & 0.667
        & 0.638
        & 0.540
        & 0.644
        & 0.578
        \\
        Self-Training
        & 0.434
        & 0.510
        & 0.497
        & 0.661
        & 0.647
        & 0.550
        & 0.672
        & 0.578
        \\
        \rowcolor{mygray}
        Ensemble
        & 0.443
        & 0.574
        & 0.627
        & 0.618
        & \textbf{0.645}
        & 0.575
        & \textbf{0.747}
        & 0.553
        \\
        \midrule
        MSL 
        & 0.446
        & 0.584
        & 0.734
        & 0.612
        & 0.632
        & 0.566
        & 0.727
        & 0.559
        \\
        DBL
        & 0.421
        & 0.536
        & 0.655
        & 0.576
        & 0.634
        & 0.581
        & 0.766
        & 0.556
        \\
        \rowcolor{mygray}
        MSL+DBL
        & \textbf{0.456}
        & \textbf{0.598}
        & \textbf{0.699}
        & \textbf{0.654}
        & \textbf{0.645}
        & \textbf{0.590}
        & 0.721
        & \textbf{0.590}
        \\
        \bottomrule
    \end{tabular}}
    \caption{\textbf{Main results.} 
    The bottom 3 rows are our results while the top 5 rows are baselines from~\cite{mapping}.
    Results in \textbf{bold} represent the better one between the baseline ensemble and our ensemble strategies.
    Compared to the baseline ensemble, which secured first place in the 2022 MICCAI ATLAS challenge, our ensemble result (MSL+DBL) consistently achieves better or equal performance on recall (\textbf{3.6\%} and \textbf{3.7\%}), F1 (\textbf{2.4\%} and \textbf{1.5\%}), and Dice scores (\textbf{1.3\%} and 0.0\%) on both the mini-lesion subset and the entire dataset, respectively.
    }
    \label{tab:result_main}
\end{table}

The evaluation of MSL, DBL, and our optimal ensemble results are illustrated in Table~\ref{tab:result_main}. To establish a solid baseline, we replicated the training of the top-performing model from the 2022 MICCAI ATLAS Challenge~\cite{mapping}. This baseline employs a standard U-Net~\cite{unet} trained with binary segmentation masks and optimized using the Dice+CrossEntropy compound loss. Building on this foundation, we implemented and assessed our proposed MSL and DBL methods. 

For evaluating small lesion detection, while we can report the small lesion detection performance of our method on the entire dataset, it is not feasible to accurately compare to baseline methods which are not explicitly size-aware. For example, these methods might segment a single large lesion, subsuming nearby small lesions that are disconnected in reality/ground truth, thus complicating the exclusion of large lesions. To address this, we selected a pure subset of 138 images exclusively containing lesions with volumes less than 1,000 voxels, denoted as the mini-lesion subset, to accurately compare the core performance of detecting small lesions.
On this subset, MSL surpasses the best ensemble results in~\cite{mapping}, which include models trained under four different schemes: the default setting, a substitution of the TopK10 loss function (DTK10), a switch to the more intricate Res U-Net architecture, and a self-training scheme leveraging an additional 300 testing scans. MSL achieves higher precision (\textbf{10.7\%}), F1 (\textbf{1.0\%}), and Dice (\textbf{0.3\%}) scores, showcasing its effectiveness in detecting small lesions.
Conversely, DBL outperforms their Default and DTK10 counterparts on both the mini-lesion subset and the entire dataset, which utilize the same network architecture. This outcome underscores the effectiveness of accentuating the boundary region.

\begin{figure*}[!t]
    \centering
    \hspace{13mm}
    \includegraphics[width=0.7\linewidth]{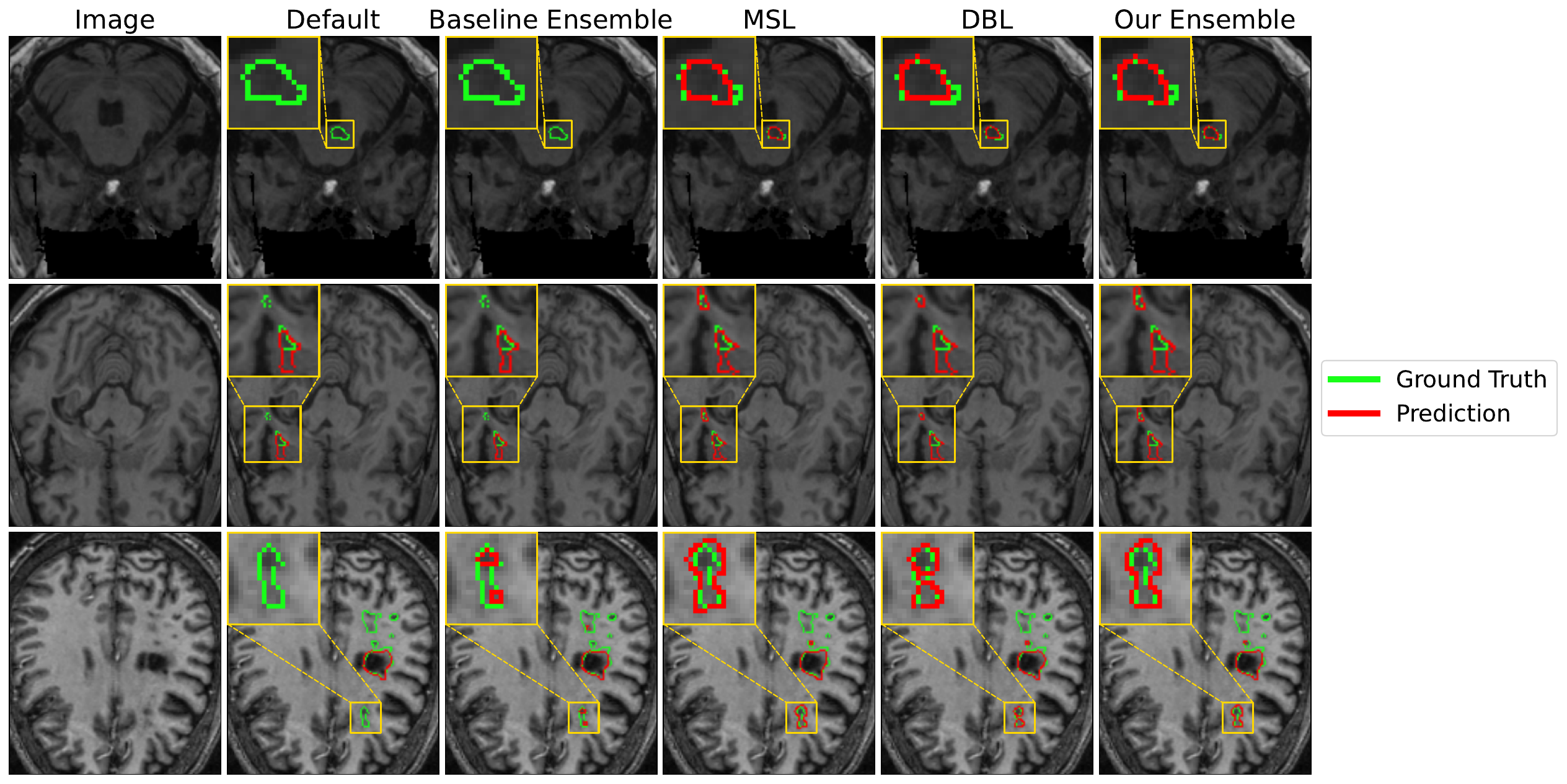}
    \caption{\textbf{Visualization of segmentation results.} 
    The green contours represent the ground truth, while the red contours depict the predicted lesions. 
    These segmentation results indicate that 
    our methods accurately label more lesions that are missed, \textit{i.e.}, false negative, by the baselines.
    }
    \label{fig:visualization}
\end{figure*}

Moreover, our optimal ensemble result which also contains four training schemes including MSL and DBL models trained on both Dice+CrossEntropy and Dice+Focal~\cite{focal} losses, consistently achieves better or equal performance on recall (\textbf{3.6\%} and \textbf{3.7\%}), F1 (\textbf{1.3\%} and \textbf{1.5\%}), and Dice scores (\textbf{1.3\%} and 0.0\%) compared to the baseline ensemble results on both the mini-lesion subset and the entire dataset, respectively.
Fig.~\ref{fig:visualization} offers qualitative segmentation results obtained from MSL, DBL, our ensemble method, and comparison with the schemes proposed in~\cite{mapping}. These results suggest our methods have a superior ability to segment small lesions.
For completeness, we have detailed the performance of MSL and DBL models trained on Dice+Focal losses, along with all post-processing hyperparameters in Sec. B and C of the supplementary materials, respectively.

\section{Ablation Studies}

\noindent
\textbf{Number of categories in MSL and DBL.} 
For MSL, we explore the separation of binary lesion masks into 3, 4, or 5 lesion classes with varying sizes, as shown in Table~\ref{tab:result_num_class}. Notably, splitting into 4 categories yields the most favorable performance, underscoring the significance of striking a balance between the advantages of isolating small lesions and the complexity associated with multi-class classification. As for DBL, we compare the outcomes of splitting solely into boundary and interior regions versus incorporating an additional transition region. Retaining only a relatively narrow boundary region achieves optimal performance, signifying the pivotal role of the lesion boundary in the segmentation process. We also provide the configuration and lesion distribution of each method in Sec. A of the supplementary materials.

\begin{table}[!t]\small
    \centering
    \scalebox{0.9}{
    \begin{tabular}{c|cccc|cccc}
        \toprule
        \multirow{2}{*}{\#Classes}
        & \multicolumn{4}{c|}{Mini-Lesion Subset}
        & \multicolumn{4}{c}{Entire Dataset}
        \\
        \cline{2-9}
        & \makebox[0.09\linewidth][c]{Dice}
        & \makebox[0.09\linewidth][c]{F1}
        & \makebox[0.09\linewidth][c]{Precision}
        & \makebox[0.09\linewidth][c]{Recall}
        & \makebox[0.09\linewidth][c]{Dice}
        & \makebox[0.09\linewidth][c]{F1}
        & \makebox[0.09\linewidth][c]{Precision}
        & \makebox[0.09\linewidth][c]{Recall}
        \\
        \midrule
        \multicolumn{9}{c}{Multi-Scale Labeling (MSL)}
        \\
        \midrule
        3
        & 0.430
        & 0.545
        & \textbf{0.749}
        & 0.553
        & 0.628
        & \textbf{0.571}
        & \textbf{0.781}
        & 0.537
        \\
        \rowcolor{mygray}
        4
        & \textbf{0.446}
        & \textbf{0.584}
        & 0.734
        & \textbf{0.612}
        & \textbf{0.632}
        & 0.566
        & 0.727
        & \textbf{0.559}
        \\
        5
        & 0.353
        & 0.463
        & 0.699
        & 0.515
        & 0.605
        & 0.522
        & 0.725
        & 0.523
        \\
        \midrule
        \multicolumn{9}{c}{Distance-Based labeling (DBL)}
        \\
        \midrule
        1
        & 0.415
        & \textbf{0.544}
        & \textbf{0.681}
        & 0.574
        & 0.632
        & 0.577
        & \textbf{0.813}
        & 0.531
        \\
        \rowcolor{mygray}
        2
        & \textbf{0.421}
        & 0.536
        & 0.655
        & \textbf{0.576}
        & \textbf{0.634}
        & \textbf{0.581}
        & 0.766
        & \textbf{0.556}
        \\
        3
        & 0.406
        & 0.525
        & 0.679
        & 0.563
        & 0.624
        & 0.574
        & 0.773
        & 0.548
        \\
        \bottomrule
    \end{tabular}}
    \caption{\textbf{Ablation studies on the number of categories for each labeling strategy.} Categorizing the binary segmentation masks into lesions with 4 different sizes yields the best MSL performance while separating into only the boundary and interior regions achieved the best DBL performance.}
    \label{tab:result_num_class}
\end{table}

\begin{figure*}[!t]
    \centering
    \begin{subfigure}[t]{\textwidth}
        \centering
        \includegraphics[scale=0.31]{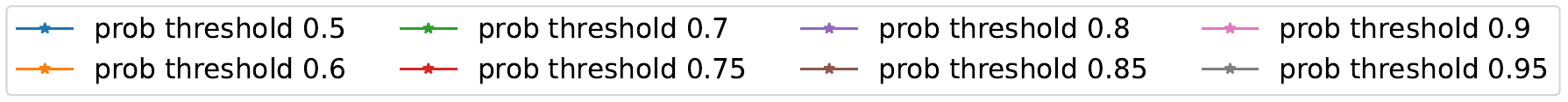}
    \end{subfigure}%
    \\
    \begin{subfigure}[t]{0.48\textwidth}
    \begin{subfigure}[t]{0.5\textwidth}
        \centering
        \hspace{-5mm}
        \includegraphics[scale=0.31]{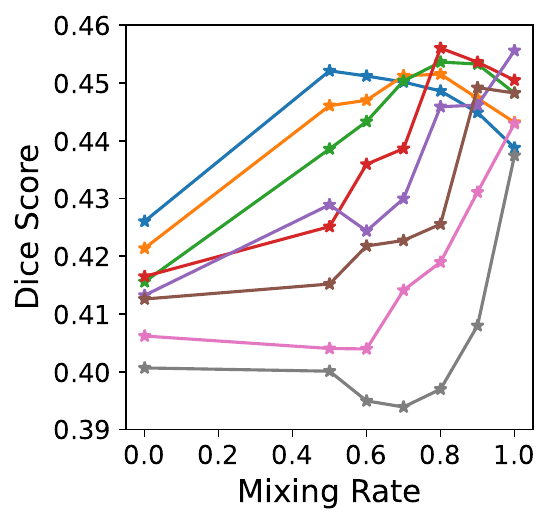}
    \end{subfigure}%
    ~ 
    \begin{subfigure}[t]{0.5\textwidth}
        \centering
        \hspace{-5mm}
        \includegraphics[scale=0.31]{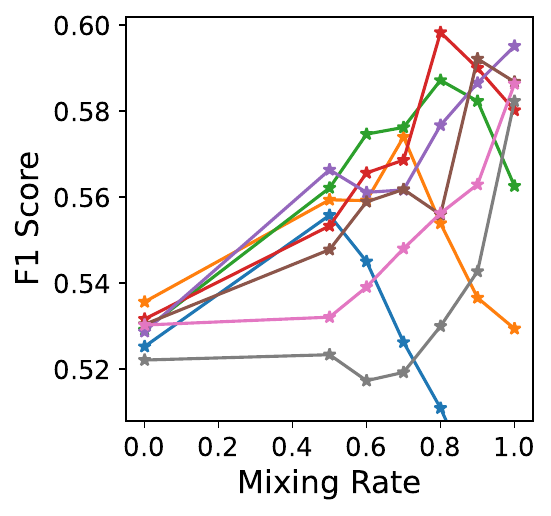}
    \end{subfigure}%
    \caption{Dice and F1 on the mini-lesion subset.}
    \end{subfigure}%
    ~
    \begin{subfigure}[t]{0.48\textwidth}
    \begin{subfigure}[t]{0.5\textwidth}
        \centering
        \includegraphics[scale=0.31]{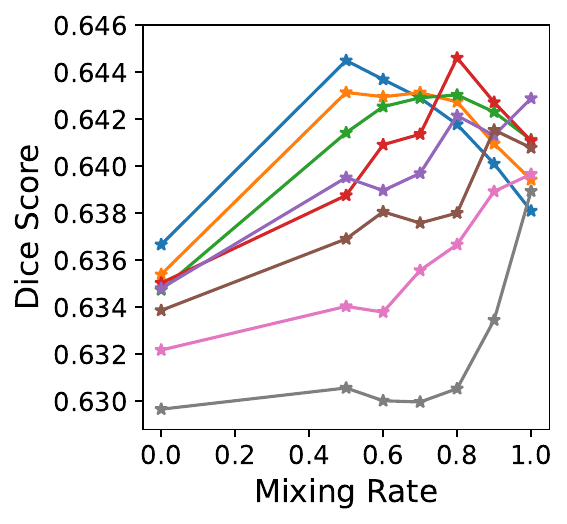}
    \end{subfigure}%
    ~ 
    \begin{subfigure}[t]{0.5\textwidth}
        \centering
        \includegraphics[scale=0.31]{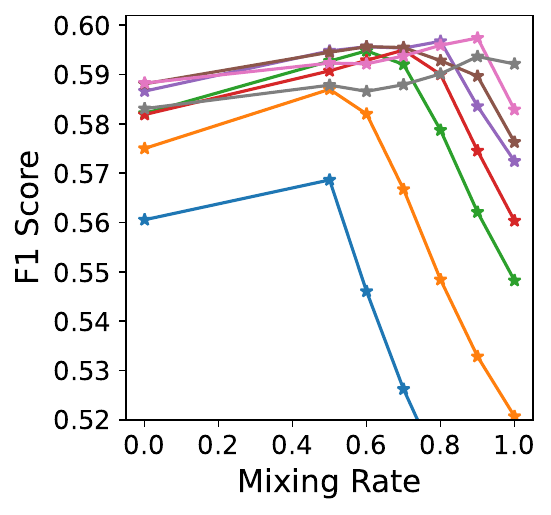}
    \end{subfigure}%
    \caption{Dice and F1 on the entire dataset.}
    \end{subfigure}%
    \caption{\textbf{Ablation studies of mixing rate and postprocessing threshold.} For Dice and F1 scores, higher is better. 
    A mixing rate of 0.8 with a post-processing threshold of 0.75 is found as the optimal configuration.
    }
    \label{fig:ablation}
\end{figure*}

\noindent
\textbf{Mixing rate and postprocessing (PP) threshold in ensemble.} 
Our ensemble and postprocessing (PP) strategies require pre-set hyperparameters, including probability thresholds and mixing rates. 
Fig.~\ref{fig:ablation} illustrates the Dice and F1 scores corresponding to various PP thresholds and mixing rates, evaluated on both the mini-lesion subset and the entire dataset. Among these configurations, a mixing rate of 0.8 paired with a PP threshold of 0.75 yields the most favorable average ranking, thus establishing our optimal setting. Notably, the Dice score shows consistent improvement with an increasing mixing rate for PP thresholds >0.6, highlighting the advantage of relying more on MSL for segmenting small lesions. The F1 score exhibits consistent behavior as well: it improves until a mixing rate of 0.5, then declines when evaluated across the entire dataset but continues to improve for the mini-lesion subset for PP thresholds >0.6.

\section{Conclusion}

This study introduces Multi-Size Labeling (MSL) and Distance-Based Labeling (DBL) methodologies, compatible with a variety of segmentation networks, aimed at enhancing the segmentation accuracy of small stroke lesions by emphasizing small lesions and the boundaries of the lesions, respectively. In tackling this challenge, our proposed ensemble strategy consistently achieves better or equal performance on recall (3.6\% and 3.7\%), F1 (2.4\% and 1.5\%), and Dice scores (1.3\% and 0.0\%) compared to the ensemble result from the top-1 winner of the 2022 MICCAI ATLAS Challenge on both the mini-lesion subset and the entire dataset, respectively. Furthermore, a single MSL model outperforms the baseline ensemble strategy with improvements of 1.0\% and 0.3\% on F1 and Dice scores, respectively. These findings underscore the effectiveness of MSL and DBL in stroke lesion segmentation, particularly in small stroke lesion segmentation.

\begin{credits}
\subsubsection{\ackname}
We would like to acknowledge the support from NIH grants R01NS123378, P50HD105353, NIH R01NS105646, NIH R01NS11102, and R01NS117568.

\subsubsection{\discintname}
The authors have no competing interests to declare that are
relevant to the content of this article. 
\end{credits}

%
%
%
\bibliographystyle{splncs04}
\bibliography{ref}
%






\appendix

\section{Lesion Distribution of MSL and DBL}
\label{sec:lesion_dist}

\begin{figure*}[!h]
    \centering
    \vspace{-9mm}
    \begin{subfigure}[!h]{0.67\textwidth}
        \centering
        \hspace{-8mm}
        \includegraphics[scale=0.25]{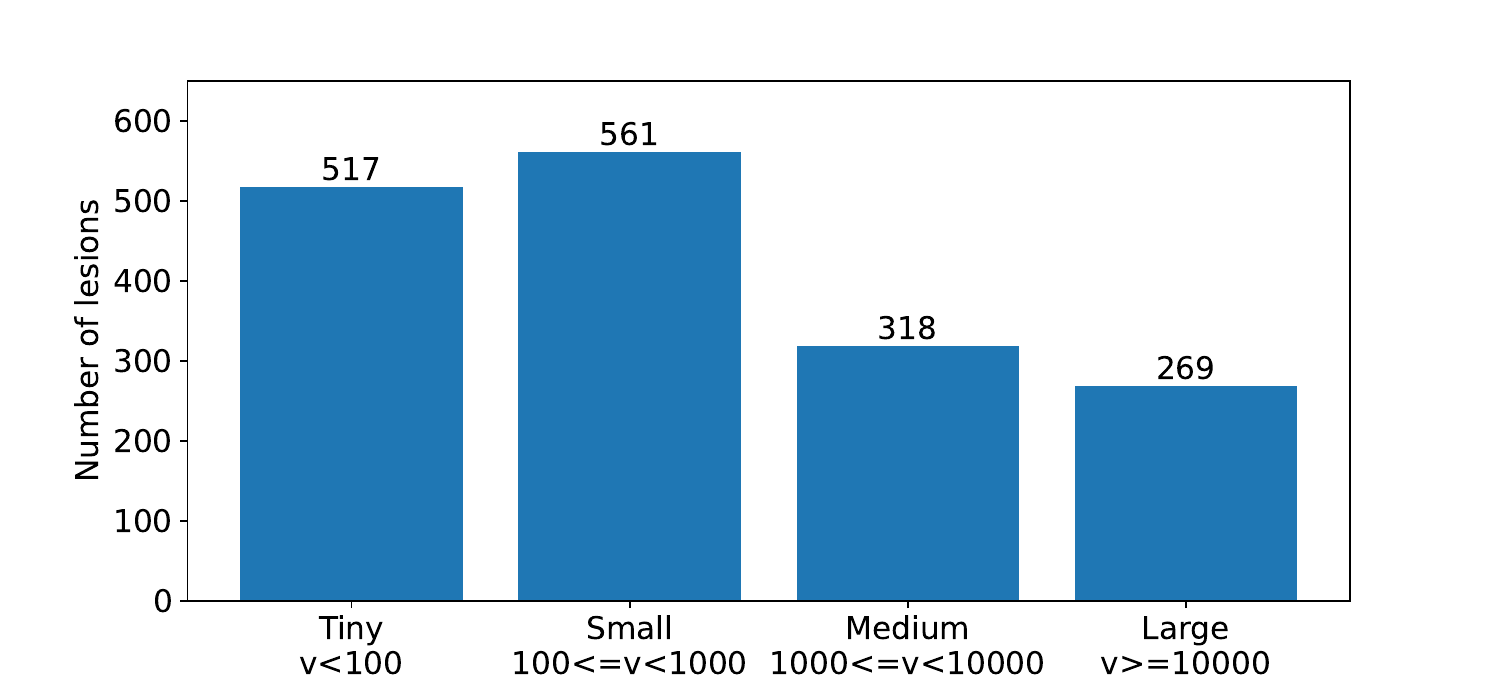}
        \caption{}
        \label{fig:msl}
    \end{subfigure}%
    ~ 
    \begin{subfigure}[!h]{0.33\textwidth}
        \centering
        \hspace{-4mm}
        \includegraphics[scale=0.25]{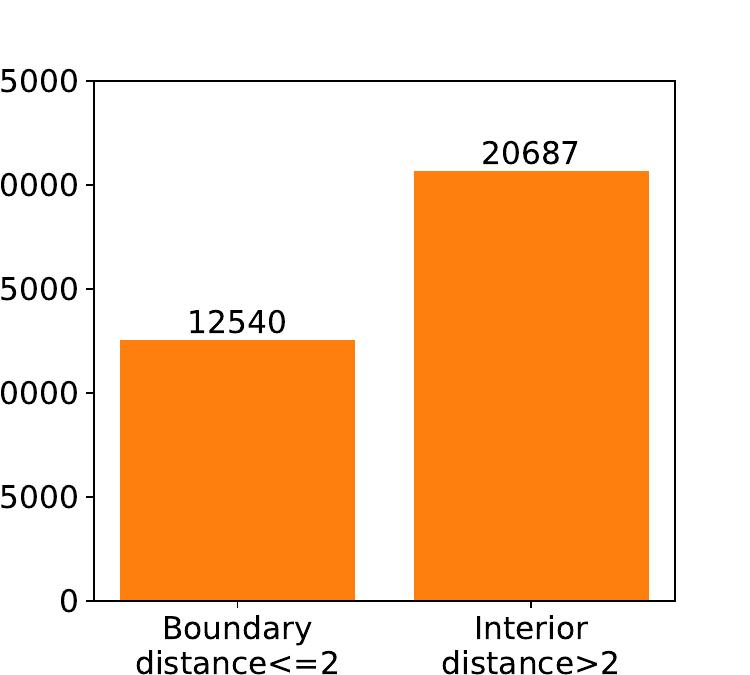}
        \caption{}
        \label{fig:dbl}
    \end{subfigure}%
    \caption{\textbf{(a) Lesion distribution of MSL categories over 655 scans in the ATLAS v2.0 dataset.} 
    The volume thresholds of each category are listed on the horizontal axis. 
    \textbf{(b) Average lesion voxels of DBL categories in each training scan in the ATLAS v2.0 dataset.} 
    Lesion voxels with a distance not larger than 2 to the non-lesion region are defined as the boundary region.
    }
    \label{fig:msl_dbl}
    \vspace{-3mm}
\end{figure*}


\begin{figure*}[!h]
    \centering
    \vspace{-9mm}
    \begin{subfigure}[!h]{0.67\textwidth}
        \centering
        \hspace{-8mm}
        \includegraphics[scale=0.25]{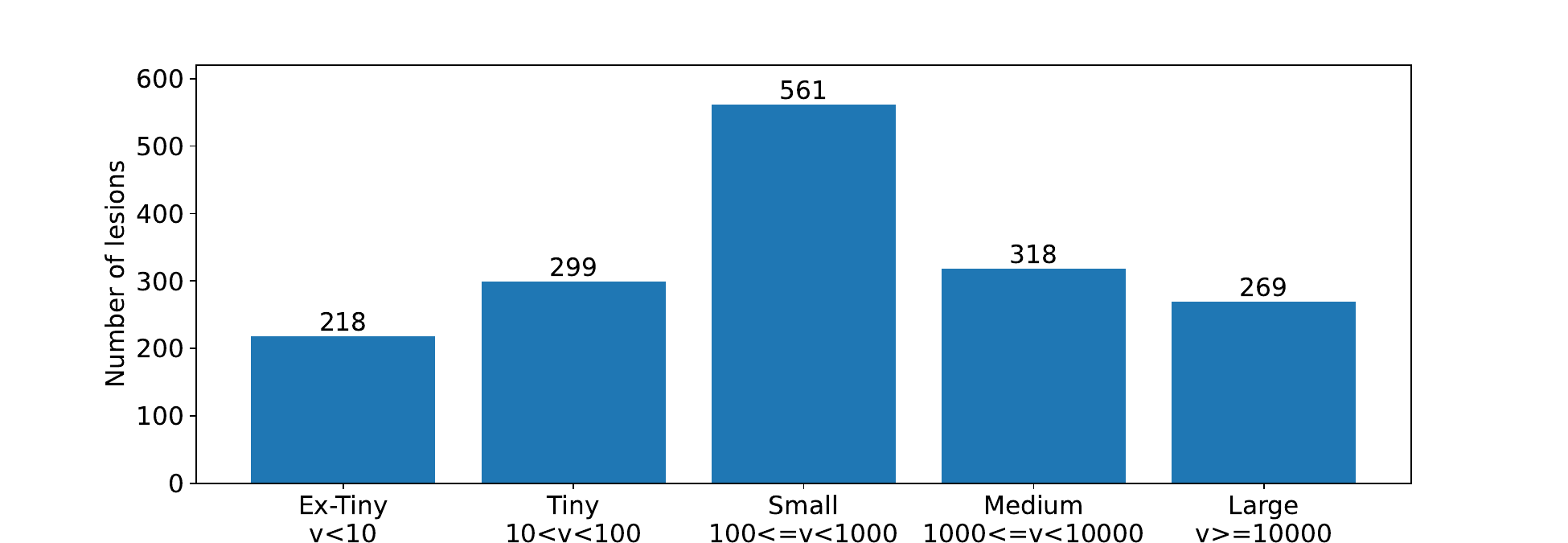}
        \caption{}
    \end{subfigure}%
    ~ 
    \begin{subfigure}[!h]{0.33\textwidth}
        \centering
        \hspace{-4mm}
        \includegraphics[scale=0.25]{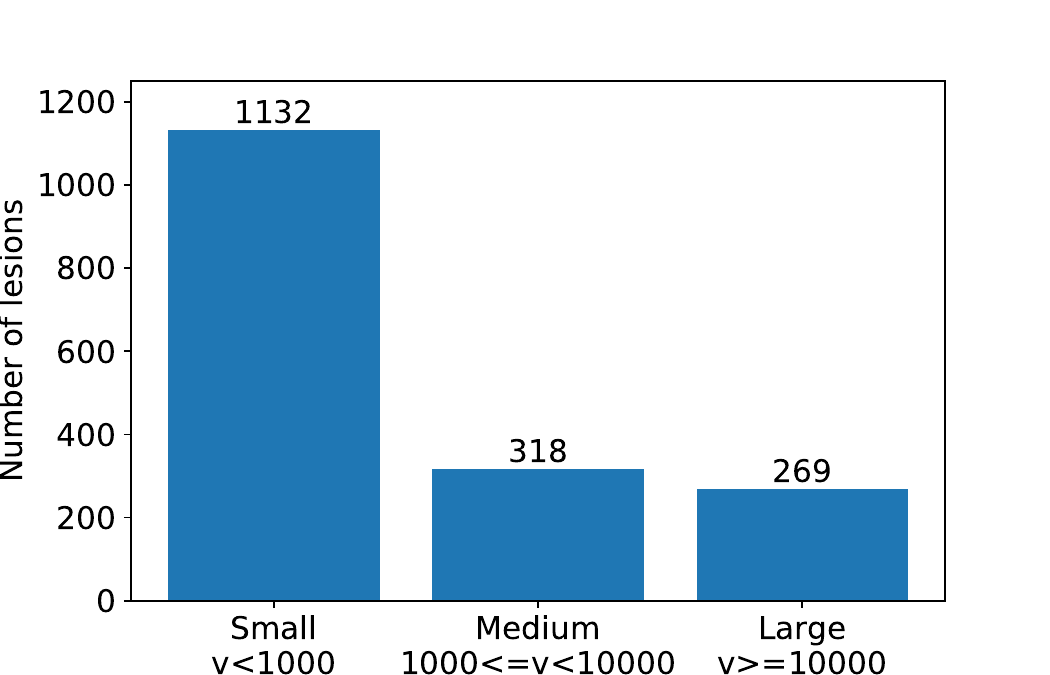}
        \caption{}
    \end{subfigure}%
    \caption{\textbf{Lesion distribution of MSL with (a) 5 categories and (b) 3 categories over 655 scans in the ATLAS v2.0 dataset.} 
    The volume thresholds of each category are listed on the horizontal axis. 
    }
    \label{fig:msl_ablation}
    \vspace{-3mm}
\end{figure*}

\begin{figure*}[!h]
    \centering
    \vspace{-9mm}
    \includegraphics[scale=0.25]{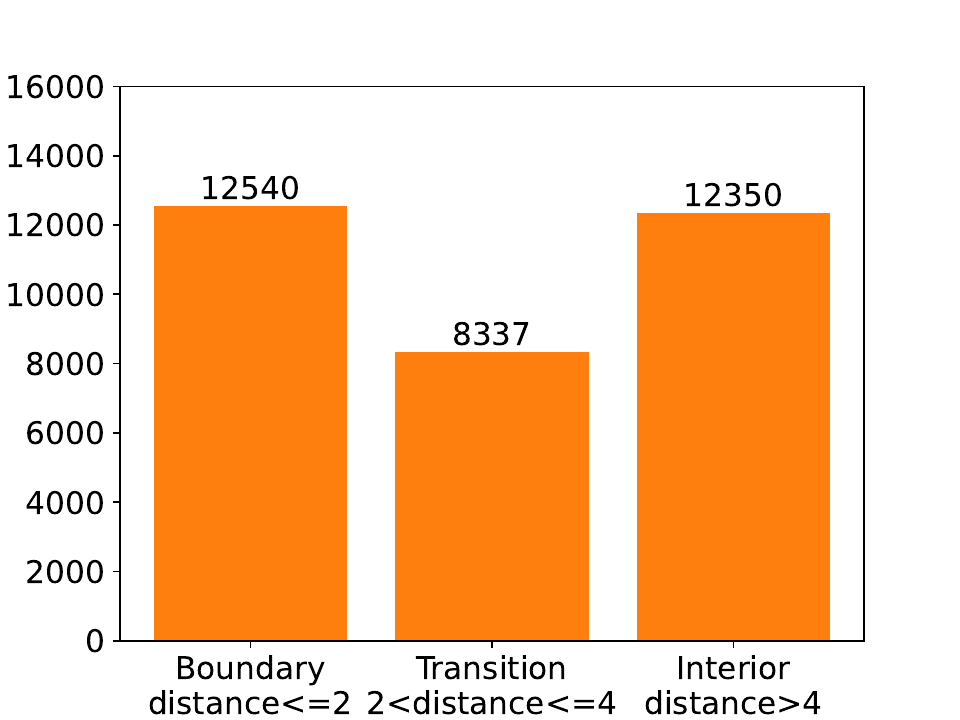}
    \caption{\textbf{Average lesion voxels of alternative DBL setting in each training scan.} 
    Lesion voxels with a distance not larger than 2 to the non-lesion region are categorized as the boundary region, while those with a distance between 2 and 4 are designated as the transition region.
    }
    \label{fig:dbl_ablation}
    \vspace{-3mm}
\end{figure*}

\section{Results on Dice+Focal Loss}
\label{sec:dice+focal}

\begin{table}[!h]
    \centering
    \begin{tabular}{l|cccc|cccc}
        \toprule
        \multirow{2}{*}{Method}
        & \multicolumn{4}{c|}{Mini-Lesion Subset}
        & \multicolumn{4}{c}{Entire Dataset}
        \\
        \cline{2-9}
        & \makebox[0.09\linewidth][c]{Dice}
        & \makebox[0.09\linewidth][c]{F1}
        & \makebox[0.09\linewidth][c]{Precision}
        & \makebox[0.09\linewidth][c]{Recall}
        & \makebox[0.09\linewidth][c]{Dice}
        & \makebox[0.09\linewidth][c]{F1}
        & \makebox[0.09\linewidth][c]{Precision}
        & \makebox[0.09\linewidth][c]{Recall}
        \\
        \midrule
        Default
        & 0.431
        & 0.527
        & 0.538
        & 0.638
        & 0.633
        & 0.557
        & 0.693
        & 0.565
        \\
        MSL
        & \textbf{0.438}
        & \textbf{0.580}
        & 0.678
        & \textbf{0.644}
        & 0.629
        & 0.560
        & 0.692
        & \textbf{0.570}
        \\
        DBL
        & 0.422
        & 0.538
        & \textbf{0.690}
        & 0.567
        & \textbf{0.635}
        & \textbf{0.588}
        & \textbf{0.791}
        & 0.551
        \\
        \bottomrule
    \end{tabular}
    \caption{\textbf{MSL and DBL results with Dice+Focal loss.}
    When the Dice+Focal loss is applied, MSL consistently outperforms the default method in the mini-lesion subset, while DBL achieves the best performance on the entire dataset.
    }
    \label{tab:result_focal}
\end{table}

\section{Postprocessing Threshold for Vanilla MSL and DBL}
\label{sec:additional_ablation}



\begin{table}[!h]
    \centering
    \begin{tabular}{ccccc}
        \toprule
        Labeling 
        & MSL
        & MSL
        & DBL
        & DBL
        \\
        \midrule
        Loss
        & Dice+CE
        & Dice+Focal
        & Dice+CE
        & Dice+Focal
        \\
        \midrule
        Postprocessing Threshold
        & 0.95
        & 0.9
        & 0.7
        & 0.8
        \\
        \bottomrule
    \end{tabular}
    \caption{\textbf{Optimal postprocessing thresholds for vanilla MSL and DBL.} 
    The optimal thresholds are determined based on the average ranking on the Dice and F1 scores across both the mini-lesion subset and the entire dataset.
    }
    \label{tab:ablation_vanilla}
\end{table}

\begin{figure*}[!h]
    \centering
    \vspace{-8mm}
    \begin{subfigure}[t]{\textwidth}
        \centering
        \includegraphics[scale=0.33]{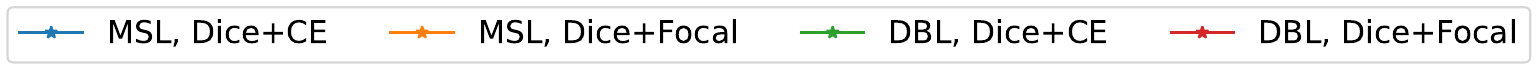}
    \end{subfigure}%
    \\
    \begin{subfigure}[t]{0.48\textwidth}
    \begin{subfigure}[t]{0.5\textwidth}
        \centering
        \hspace{-5mm}
        \includegraphics[scale=0.33]{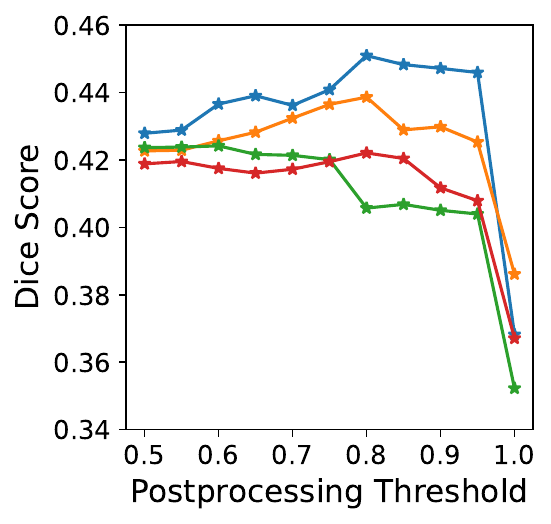}
    \end{subfigure}%
    ~ 
    \begin{subfigure}[t]{0.5\textwidth}
        \centering
        \hspace{-5mm}
        \includegraphics[scale=0.33]{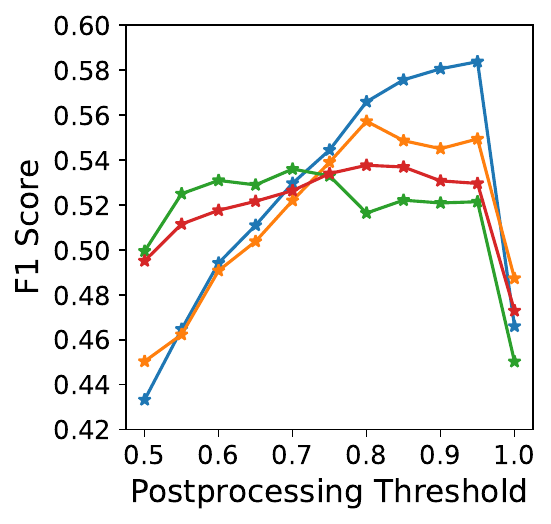}
    \end{subfigure}%
    \caption{Dice and F1 on the mini-lesion subset.}
    \end{subfigure}%
    ~
    \begin{subfigure}[t]{0.48\textwidth}
    \begin{subfigure}[t]{0.5\textwidth}
        \centering
        \includegraphics[scale=0.33]{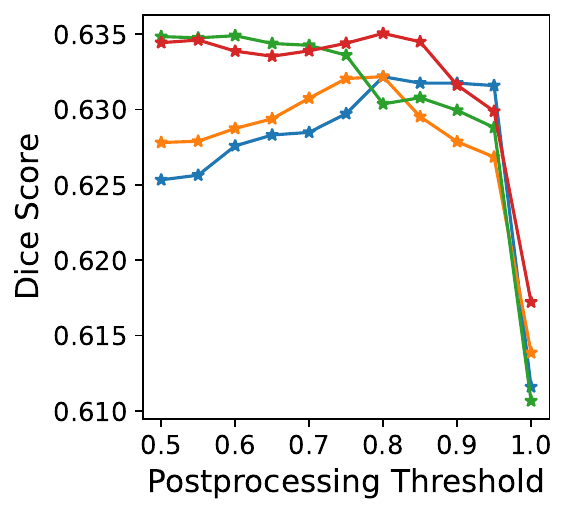}
    \end{subfigure}%
    ~ 
    \begin{subfigure}[t]{0.5\textwidth}
        \centering
        \includegraphics[scale=0.33]{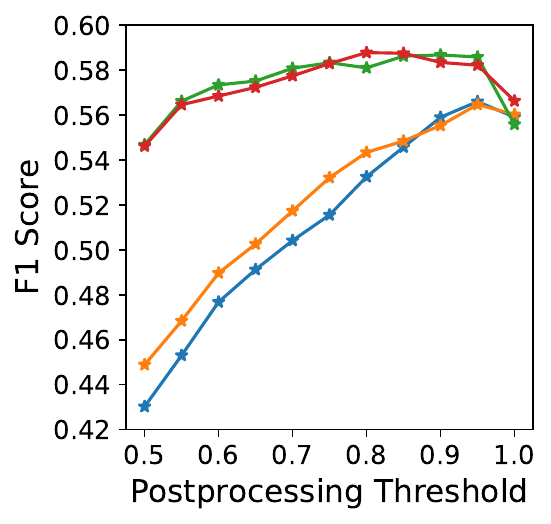}
    \end{subfigure}%
    \caption{Dice and F1 on the entire dataset.}
    \end{subfigure}%
    \caption{\textbf{Ablation studies of postprocessing thresholds on vanilla MSL and DBL.} 
    The optimal postprocessing thresholds for each method, determined based on the average ranking on the Dice and F1 scores, are presented in Table~\ref{tab:ablation_vanilla}.
    }
    \label{fig:ablation_vanilla}
\end{figure*}

\end{document}